\UseRawInputEncoding
\def\year{2023}\relax
\documentclass[letterpaper]{article} % DO NOT CHANGE THIS
\usepackage{aaai23}  % DO NOT CHANGE THIS
\usepackage{times}  % DO NOT CHANGE THIS
\usepackage{helvet}  % DO NOT CHANGE THIS
\usepackage{courier}  % DO NOT CHANGE THIS
\usepackage[hyphens]{url}  % DO NOT CHANGE THIS
\usepackage{graphicx} % DO NOT CHANGE THIS
\usepackage{natbib}  % DO NOT CHANGE THIS AND DO NOT ADD ANY OPTIONS TO IT
\usepackage{caption} % DO NOT CHANGE THIS AND DO NOT ADD ANY OPTIONS TO IT
\DeclareCaptionStyle{ruled}{labelfont=normalfont,labelsep=colon,strut=off} % DO NOT CHANGE THIS
\usepackage{algorithm}
\usepackage{algorithmic}

\usepackage[table,xcdraw]{xcolor} % FW edited
\definecolor{comments}{rgb}{0.6, 0.2, 0.9}  % only used in pseudo code
\usepackage{amsmath} % FW added
\usepackage{verbatim}  % FW added
\usepackage{fancyhdr}
\usepackage{newfloat}
\usepackage{listings}

\DeclareCaptionStyle{ruled}{labelfont=normalfont,labelsep=colon,strut=off} % DO NOT CHANGE THIS
\floatstyle{ruled}
\newfloat{listing}{tb}{lst}{}
\floatname{listing}{Listing}

\title{MuMIC - Multimodal Embedding for Multi-label Image Classification with Tempered Sigmoid}

\author{
  Fengjun Wang,
  Sarai Mizrachi,
  Moran Beladev, 
  Guy Nadav, \\
  Gil Amsalem,
  Karen Lastmann Assaraf,
  Hadas Harush Boker
  }

\affiliations{
    Booking.com\\
    
    \{fengjun.wang, sarai.mizrachi, moran.beladev, guy.nadav, gil.amsalem, karen.lastmannassaraf, hadas.harush\}@booking.com
}

\begin{document}
\thispagestyle{fancy}
\chead{This paper was accepted to IAAI 2023. Please reference it instead once published}
\rhead{} %empt

\maketitle
\begin{abstract}
% \guyn{Multi-label image classification is a foundational topic in many domains. A multimodal learning approach has recently achieved outstanding results in image representation and single-label image classification. As an example, Contrastive Language Image Pretraining (CLIP) demonstrated impressive learning abilities and is robust to natural distribution shifts. This success inspires us to leverage multimodal learning for multi-label classification tasks.  In this study, a total of 120 image classes are defined, and more than 140K positive annotations are collected on approximately 60K Booking.com images. We propose a Multimodal Multi-label Image Classification (MuMIC) framework which is capable of providing high classification performance on defined classes, supporting zero-shot prediction, and producing domain-specific image embeddings. This paper introduces a modification to the CLIP model that utilizes tempered sigmoid based binary cross entropy loss in order to handle multi-label classification. Our modeling choices are extensively tested through ablation studies. Our final deployed model outperforms other state-of-the-art models with 85.6 GAP@10 score, 83.8 GAP score on all 120 classes, and 90.1 macro mAP score across 32 majority classes. To the best of our knowledge, we are the first to apply contrastively learned multimodal pretraining to real-world multilabel classification problems.}

Multi-label image classification is a foundational topic in various domains. 
% Besides popular convolutional neural networks (CNNs) based and transformer based architectures, 
Multimodal learning approaches have recently achieved outstanding results in image representation and single-label image classification. For instance, Contrastive Language-Image Pretraining (CLIP) demonstrates impressive image-text representation learning abilities 
and is robust to natural distribution shifts. 
This success inspires us to leverage multimodal learning for 
multi-label classification tasks, and benefit from contrastively learnt pretrained models. 

We propose the Multimodal Multi-label Image Classification (MuMIC) framework, which utilizes a hardness-aware tempered sigmoid based Binary Cross Entropy loss function, thus enables the optimization on multi-label objectives and transfer learning on CLIP. 
MuMIC is capable of providing high classification performance, handling real-world noisy data, supporting zero-shot predictions, 
and producing domain-specific image embeddings. % travel domain specific?

In this study, a total of 120 image classes are defined, and more than 140K positive annotations
are collected on approximately 60K Booking.com images. The final MuMIC model is deployed on Booking.com Content Intelligence Platform, and it outperforms other state-of-the-art models with $85.6\%$ GAP@10 and $83.8\%$ GAP on all 120 classes, as well as a $90.1\%$ macro mAP score across 32 majority classes. We summarize the modelling choices which are extensively tested through ablation studies. To the best of our knowledge, we are the first to adapt contrastively learnt multimodal pretraining for real-world multi-label image classification problems, and the innovation can be transferred to other domains. 
% We discuss the development, deployment and use cases.  
% where the lessons learnt can be transferred to other domains. 
% We present the system’s architectural design in production and discuss its use and impact.

% \\MuMIC introduces modifications to the CLIP model, mainly by utilizing tempered sigmoid based binary cross entropy loss, which enables the optimization on multi-label objectives, and provides hardness-aware property. 
% \monia{We propose a novel approach for multi-label classification leveraging multimodal input. Particularly, we introduce a modification of the CLIP model to handle multi-label image classification by applying tempered sigmoid. }
% Extensive experiments describe ablation results during model tuning and show decisions on - temperature factor choice; prompt descriptions on a subset of classes; compare with baselines.
 
% \monia{We conduct extensive ablation studies on our adaptation choices. } 

\end{abstract}

\section{Introduction}
\label{sec:intro}

Multi-label image classification has been widely studied with supervised learning approaches, from various Convolutional Neural Networks (CNNs) based models \cite{o2015introduction} to Vision Transformers  \cite{dosovitskiy2020image}. 
Transfer learning on pretrained models becomes the first choice for many domain-specific applications. 
% , by fine-tuning pretrained models on downstream tasks. -> redundant. I remove. 

At Booking.com, there are more than 350 million images from multiple sources. 
% , where images have shown to be a core information help travellers make decision on booking an hotel. -> redundant 
Understanding image content is crucial for both travellers and property owners, which makes image classification a core component serving as the backbone of many applications, as described in Section \ref{sec:application}. 

To drive the various use cases, we need the ability to classify images to a large set of possible labels at a massive scale. 
The first step is to formulate a label definition list to cover the most important classes. 
In Section \ref{sec:dataset}, we describe the label definition, label hierarchy, and the annotation procedure. 
% question designs with smart grouping strategies. 
Besides the predefined labels, new travel trends would probably require us to predict on unseen classes. 
% Thus, we prefer to build a solution with zero-shot learning abilities. 
Thus, we need an efficient and scalable solution with zero-shot learning abilities. 
% \cite{zero_shot_good_bad}. 
% Thus, we aim to provide a reliable solution which is also scalable to zero-shot \cite{zero_shot_good_bad}. % TODO: is this `aka` good? 
% Thus, we aim to provide a solution which is reliable, measurable, and also scalable to zero-shot prediction. % TODO: is this `aka` good? 

% Compared with single-label classification, 
We first explore state-of-the-art (SOTA) multi-label classification methods which rank high in open image datasets. They typically handle two main challenges: label imbalance, and extracting features from different regions for multiple objects. 
% The multi-label image classification has two main challenges: label imbalance, and extracting features from different regions for multiple objects.  
% global-local attention (capture the whole scene and regions for multiple objects).
To handle the label imbalance, \citet{ridnik2021asymmetric} introduces Asymmetric Loss (ASL) that down-weights and hard-thresholds easy negative samples, while also decouples the penalties on misclassifying the positive and negative samples, which is a novel improvement compared to Focal Loss \cite{focal_loss}. 
% Besides the loss function itself, ASL provides pretrained TResNet \cite{tresnet} based models which rank high in various open image datasets. Although ASL doesn't support zero-shot, we still fine-tune their pretrained models as one baseline.  -> easy to confuse readers. we use pretrained models from ASL loss, with tresnet. 
For the second challenge, many methods are proposed to learn semantic label embeddings and attentions to visual features, like cross-modality attention \cite{You2020CrossModalityAW}, and Query2Label \cite{query2label}.  
% rank high in open image datasets recently. 
% become SOTA model recently.  transformer based approaches
% and rank high in open image datasets. 
However, all the above approaches do not show zero-shot abilities.  
ML-decoder \cite{ridnik2021ml} is another SOTA method, which proposes a new classifier head supporting efficient training on large number of classes, and can generalize to unseen classes. 
However, the zero-shot setting requires specific group-decoding designs, and uses one shared projection matrix in the group fully-connected layer for all classes, which involves additional model training and might cause performance drop. Hence, we do observe the gaps between the current SOTA methods and our requirements. 
% dynamically down-weights and hard-thresholds easy negative samples, while also discarding possibly mislabeled samples. They demonstrate how ASL can balance the probabilities of different samples 

% Image representation learning is another popular task that does not solve the classification problem directly, but provides useful image embeddings for downstream tasks. 
We further explore image representation learning works and aim to find zero-shot potential and robust pretrained models. 
% , which could provide supports for downstream tasks like image classification. 
It has recently been shown that contrastive representation learning on images is superior to learning from equivalent predictive objectives \cite{cl_img_obj2019}. 
One specific approach is using natural language processing (NLP) supervision on images to contrastively enforce better learning of visual concepts \cite{nlp_contras_zhang_2020}. 
NLP supervision enables zero-shot transfer ability to generalize to unseen image-text matching patterns, and can leverage on broad image-text datasets. 

CLIP \cite{radford2021learning} is a top performer that uses such NLP supervision, trained on 400 million image-text pairs, and provides robust as well as efficient image representations. 
CLIP trains multiple image encoding backbones - 5 ResNets \cite{resnet_2016} and 3 Vision Transformers (ViT) \cite{dosovitskiy2020image}, and shows that ViT provides more computational efficiency which allows production of high performance models. 
As for image classification, CLIP mainly shows the performance on single-label tasks.
To demonstrate transfer learning ability, CLIP focuses on benchmarking zero-shot performance based on image-text embedding similarity, and few-shot linear classifiers. It does not demonstrate fine-tuned results but indicates high potential on it. 

Our primary focus is on training visual and text transformers in an image-text pairs pattern with a modified loss function for multi-label classification.
The paper's contributions can be summarized as follow:  

\begin{itemize}
    \item We design a practical MuMIC framework 
    by introducing a hardness-aware loss function - tempered sigmoid based Binary Cross Entropy (BCE) - for multi-label classification, and leveraging on transfer learning from CLIP.  
    % on the travel domain. -> MuMIC framework is not only for travel. We conclude travel domain contribution in Conclusion
    \item We share the core findings during the model development, including: theoretically how the sigmoid temperature controls the strengths of penalties on hard samples, as well as how to tune it practically; 
    enrichment on class context (label-text) representation; 
    domain specific image preprocessing etc.
    \item We study and share efficient and practical dataset annotation strategies, as described in Section \ref{sec:dataset}. 
    \item We compare the final best model's performance to two baselines: ASL and CLIP. 
    % the original CLIP. 
    We visualize the improved MuMIC image embedding and compare it with CLIP. 
    \item We provide zero-shot capability for unseen classes, which performs better than CLIP when the unseen class is in the travel domain, and improves product scalability. 
    \item 
    % We deploy the final product at Booking.com and share the findings. 
    We deploy the final product at Booking.com, and share the main deployment findings. 
    % As a practical framework, MuMIC can be applied to other domains. 
\end{itemize}

\section{Approach}  % Method
\label{sec:approach}

% Given a batch of $N$ images, we have $(I, T)$ pairs, where 
% $I$ is a preprocessed image and $T$ is a vector of tokenized label-text (a query text per label), MuMIC is designed to predict which of the labels are relevant and present in the image. 
% Given a batch of $N$ images, and $L$ labels, we have $N * L$ $(I, T)$ pairs, where $I$ is an image embedding, and $T$ is a vector of tokenized label-text (one text per label), MuMIC is designed to predict which of the labels are relevant and present in the image. 

Given a batch of $N$ images, and $L$ labels, we have $N \times L$ image-label pairs. MuMIC generates $N \times L$ pairs of $(I, T)$, where $I$ is an image embedding, and $T$ is a label embedding (represents the class context). MuMIC is designed to predict which of the labels are relevant and present in the image. 

% CLIP paper said: At the core of our approach is the idea of learning image perception from supervision contained in natural language. 
% Typically, we pre-train ViT on large datasets, and fine-tune to (smaller) downstream tasks. 
Our general approach is adapting CLIP to support multi-label classification, while learning visual perceptions from natural language supervision of class context. 
Specifically, we apply ViT \cite{dosovitskiy2020image} as the computer vision backbone, and Transformer \cite{vaswani2017attention} as the text backbone; with initialized weights from pretrained CLIP, we further fine-tune on the travel domain-specific dataset. This section describes the main proposed method.  
% CLIP applied ViT \cite{dosovitskiy2020image} as image representation backbone, and Transformer \cite{vaswani2017attention} as text backbone. Since CLIP pretrained model learns from 400 million image-text pairs and has 

\begin{figure*}[ht]
 \centering
    \includegraphics[width=0.9\textwidth]{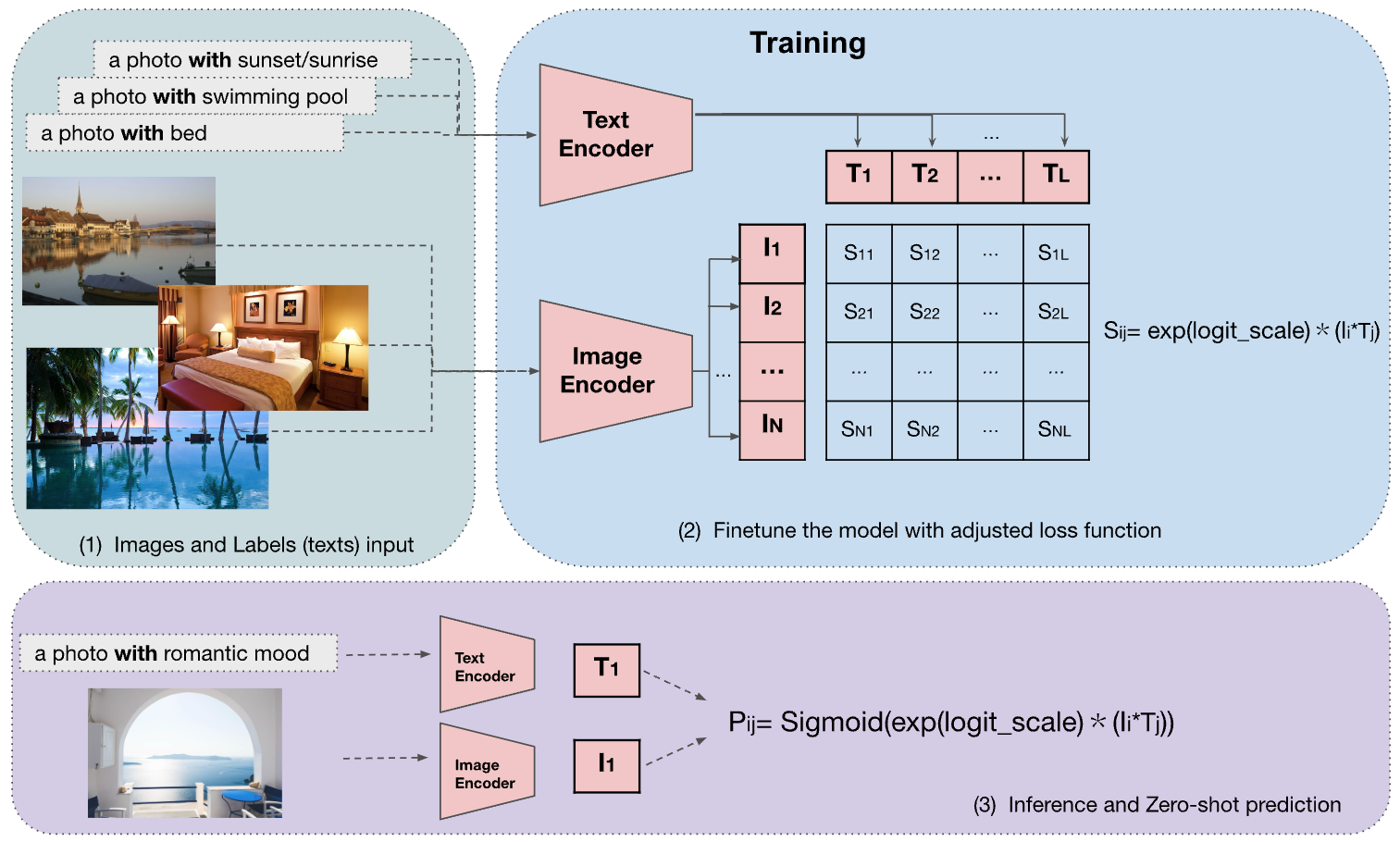}
    \caption{MuMIC architecture. (1) Each label is represented as a label-text like ``a photo with \{label name\}", or ``a photo with \{label description\}". The Image Encoder input is a batch of preprocessed images. The Text Encoder input is a list of vectors (each vector is a tokenized label-text). The $S_{ij}$ is MuMIC output logit on image $i$ and label $j$. 
    (2) For inference and zero-shot prediction, MuMIC generates predication scores ($P_{ij}$) for image and label-text by - apply temperature scaling on (image, text) embeddings' cosine similarity, and then apply sigmoid function on the scaled cosine similarity.}
    \label{fig:model_architecture}
\end{figure*} 

\subsection{Main Framework} 
\label{sec:main_framework}
% https://tex.stackexchange.com/questions/545077/color-of-python-comment-keywords-in-latex 

% \guyn{MuMIC learns a multimodal embedding space by jointly training an image encoder and text encoder to maximize the cosine similarity of the image and text embeddings of real labels, as shown in Figure  \ref{fig:model_architecture}. We apply tempered sigmoid based BCE loss on each class separately and then mean-reduce it (see equation \ref{'f:lnc'}), optimizing across all classes, using the cosine similarity of each (image, label-text) combination.} 
MuMIC learns a multimodal embedding space by jointly training an image encoder and text encoder to maximize the cosine similarity of the image and label-text embeddings of real labels, as shown in Figure  \ref{fig:model_architecture}. We apply tempered sigmoid based BCE loss on each class and then mean-reduce it (see equation \ref{'f:lnc'}), optimizing across all classes. 
% using the cosine similarity of each (image, label-text) combination.

The Listing \ref{lst:listing} describes MuMIC's core implementation, and demonstrates changes on top of CLIP. Instead of getting image-caption pairs as input, we input a batch of images, together with a list of tokenized texts where each text is derived from either the class name, or the class description. 

% \begin{figure}[H]
%     \centering
%     \includegraphics[width=0.9\columnwidth]{./figures/pseudocode.png}
%         \caption{Numpy-like pseudocode for MuMIC core implementation}
%     \label{fig:code}
% \end{figure} 

\begin{listing}[tb]%
\caption{Numpy style pseudocode {\tt mumic\_core.py}}%
\label{lst:listing}%
\begin{lstlisting}[language=Python]
# image_encoder - Vision Transformer
# text_encoder  - Text Transformer
# I[n, h, w, c] - minibatch of aligned images
# T[n_c, l]     - vector of tokenized label-text (n_c is #classes, l is #tokens)
# W_i[d_i, d_e] - learned image proj
# W_t[d_t, d_e] - learned text proj
# targets[n, n_c]  - the ground truth
# logit_scale   - the logit_scale param, = ln(1/temperature)

# extract feature representations of each modality
I_f = image_encoder(I) #[n, d_i]
T_f = text_encoder(T)  #[n_c, d_t]

# joint multimodal embedding 
I_e = l2_normalize(np.dot(I_f, W_i), axis=1)  # [n, d_e]
T_e = l2_normalize(np.dot(T_f, W_t), axis=1)  # [n_c, d_e]

# scaled pairwise cos similarities [n, n_c]
logits = np.dot(I_e, T_e.T) * np.exp(logit_scale) 
# loss - only need image level loss
loss = nn.BCEWithLogitsLoss(logits, targets) 
\end{lstlisting}
\end{listing}

% \lstset{language=Python, 
%         % basicstyle=\ttfamily\small, 
%         commentstyle=\color{comments}
% }
% \begin{lstlisting}
% # image_encoder - Vision Transformer
% # text_encoder  - Text Transformer
% # I[n, h, w, c] - minibatch of aligned images
% # T[n_c, l]     - vector of tokenized label-text (n_c is #classes, l is #tokens)
% # W_i[d_i, d_e] - learned image proj
% # W_t[d_t, d_e] - learned text proj
% # targets[n, n_c]  - the ground truth
% # logit_scale   - the logit_scale param, = ln(1/temperature)

% # extract feature representations of each modality
% I_f = image_encoder(I) #[n, d_i]
% T_f = text_encoder(T)  #[n_c, d_t]

% # joint multimodal embedding 
% I_e = l2_normalize(np.dot(I_f, W_i), axis=1)  # [n, d_e]
% T_e = l2_normalize(np.dot(T_f, W_t), axis=1)  # [n_c, d_e]

% # scaled pairwise cos similarities [n, n_c]
% logits = np.dot(I_e, T_e.T) * np.exp(logit_scale) 
% # loss - only need image level loss
% loss = nn.BCEWithLogitsLoss(logits, targets) 

% \end{lstlisting}  

\subsection{Binary Cross Entropy Loss, based on Tempered Sigmoid} 
\label{sec:bce_loss}
\subsubsection{BCE Loss}

% For the classification loss, 
With each input image $i$, given the ground-truth multi-label vector $y_{i}$, 
% , given the output logits $x_{i}$ (the cosine similarity scores) and the ground-truth multi-label vector $y_{i}$, 
we apply below BCE loss function: 

\begin{equation}
\label{'f:lnc'}
\begin{aligned}
% \begin{split}
L_{BCE} = - \frac{1}{N} \frac{1}{L} \sum_{i=1}^{N} \sum_{j=1}^{L} ({ p_j y_{ij} \cdot \log \sigma(x_{ij})} \\
{+ (1 - y_{ij}) \cdot \log (1 - \sigma(x_{ij}))  } 
% \end{split}
\end{aligned}
\end{equation}

where $\sigma(\cdot)$ is the Tempered Sigmoid function; 
$x_{ij}$ is the original output logits (before applying temperature scaling), which is the pairwise image-text cosine similarity score on image $i$ and class $j$; 
and $p_j$ is the positive samples weight of class $j$. 
A higher $p_j$ indicates that positive samples are given greater weight, increasing the penalty for identifying false negatives.
% Higher $p_j$ means more weight on positive samples, so that more penalty of classifying positive examples as negative.  

As shown in line 21 of Listing \ref{lst:listing}, we combine the BCE loss with the tempered sigmoid calculation in one single layer. It is more numerically stable by taking advantage of the log-sum-exp trick, which is widely used in machine learning \cite{nielsen2016guaranteed}.

\subsubsection{Tempered Sigmoid}
\label{sec:tempered_sig}

As described in \cite{Papernot_Thakurta_Song_Chien_Erlingsson_2021}, the tempered sigmoid function family has 3 hyperparameters - scale, temperature, and offset. Considering we are only using tempered sigmoid for the output layer, we do not use the scale and offset parameters, and the formula we apply is: 
% TODO: do we add some explanation, or it is straightforward. 
% \begin{equation}
\begin{flalign}
\begin{split}
\sigma(x) 
&= \frac{1}{1 + \exp(-x/\tau) } \\
% = \frac{1}{1 + \exp(-x * ln(1/\tau))) } \\
&= \frac{1}{1 + \exp(-x \cdot \exp(logit\_scale)) } 
\end{split}
\end{flalign}  % use flalign, then don't need use equation again 
% \end{equation}

where $\tau$ is the temperature, $logit\_scale$ is the log-parameterized multiplicative scalar as mentioned in Listing \ref{lst:listing} line 8, and $x \cdot \exp(logit\_scale)$ is the output logit. 

\subsubsection{Why our BCE Loss is hardness-aware}

Tempered losses are recently found to provide robustness to noise during training \cite{DBLP:journals/corr/abs-1906-03361}. 
\citet{understand_cl} provides a theoretical proof on why temperature makes softmax-based contrastive loss hardness-aware, and balances the closeness tolerance as well as the ability to learn separable features. 
\citet{Papernot_Thakurta_Song_Chien_Erlingsson_2021} applies tempered sigmoid activations to replace unbounded activation functions like RELU, and show better performance on noisy data.

In multi-label classification problems, sigmoid-based BCE is widely used. We are inspired by the above mentioned papers, to experiment on adding temperature for sigmoid function at the last layer, and use the temperature to make BCE a hardness-aware loss function. 
% , which is practical for real-world problems. 
With the BCE loss formula, we can derive the gradient of the loss function to the original logits value $x_{ij}$ (for image $i$ and class $j$): 
\begin{equation}
\frac{\partial L_{BCE} (x_{ij}, y_{ij})}{\partial x_{ij}} = \frac{1}{\tau} (\sigma(x_{ij}) - y_{ij})
\end{equation}

% derivative = lambda x,t,y : (sigmoid(x,t)-y)*(t) 

\begin{figure}
    \centering
    \includegraphics[width=\columnwidth]{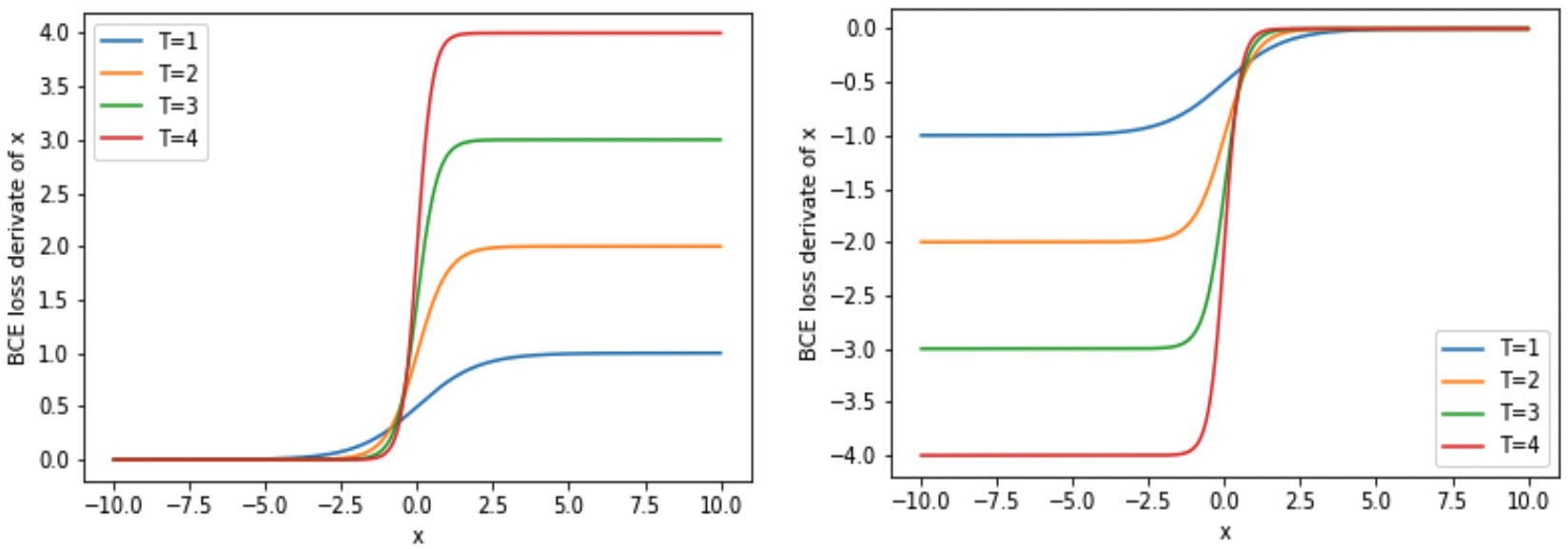}
        \caption{Our BCE loss's derivative of x. Left side: for negative samples. Right side: for positive samples.}
    \label{fig:temp_sig_2}
\end{figure}

% \begin{figure}
%     \centering
%     \includegraphics[width=0.75\columnwidth]{./figures/temp_sig_pos copy.png}
%         \caption{BCE loss's derivative of x, for positive samples. T is the inverse temperature}
%     \label{fig:temp_sig_pos}
% \end{figure} 

% \begin{figure}
%     \centering
%     \includegraphics[width=0.75\columnwidth]{./figures/temp_sig_neg copy.png}
%         \caption{BCE loss's derivative of x, for negative samples. T is the inverse temperature}
%     \label{fig:temp_sig_neg}
% \end{figure}  

In Figure \ref{fig:temp_sig_2} 
% and Figure \ref{fig:temp_sig_neg}, 
we visualize the above partial gradient function, where $T$ is the inverse temperature. From both formula and plots, it is clear that temperature plays a significant role in controlling the strength of penalties on hard samples. 
We observe that with smaller temperature (higher $T$), the derivative of loss has larger magnitude, which is proportional to the gap between the prediction probability ($\sigma(x_{ij})$) and the ground truth ($y_{ij}$). 
Thus, the loss is more sensitive to hard samples and will enforce stronger improvements on them. In Section \ref{sec:results}, we conclude the impact of different temperature levels by showing ablation experiments results, and also compare with normal BCE loss ($\tau$ as 1). 

\subsection{Image Preprocessing}
For image preprocessing, CLIP applies random square crop on resized images for training, and center square crop on resized images for inference. With our use case, it could happen that objects appear at the long edge side. For example, shower and bathtub are sometimes not the central focus of a bathroom image. The square crops could cut them out, while we'd like to recognize them. Our experimental studies show that feeding original images resized to $224 \times 224$ with no cropping yields best results as we do not lose important information about a whole scene, and so it is chosen as our final image preprocessing approach. For example, the average precision on TV/Multimedia, and Coffee facilities, are improved by 8.5\%, and 7.3\% respectively, by removing the cropping step as described above.

\section{In-house Dataset Construction}  
\label{sec:dataset}
In this section, we describe the creation of the Booking.com multi-label image annotation dataset. 

\subsection{Class Label Definition}
We create a representative label list covering the majority of potential applications. The final list has a total of 120 classes and 8 categories. For each class, we define a label name, a label description, and assign one category for it. One example of a label name is ``Historical structure", which is described as ``historical structure, a building or structure with historical value", and is assigned to the ``Outside" category. 
The 120 classes contain both tangible objects with different sizes (e.g. mirror, kitchen), and intangible items (e.g. winter, sunset). 
Besides that, there are hierarchical patterns: for example, ``Swimming pool" class is a parent of both ``Indoor swimming pool" class and ``Outdoor swimming pool" class. 

\subsection{Annotation Collection} 

\subsubsection{Image Candidate Selection}
Considering the natural class imbalance, we apply two strategies to generate image samples from the Booking.com database: 

1) Enrich image diversity: knowing which property each image belongs to, we apply stratified sampling considering the property features, for example property type (e.g. Villa).
2) Enrich low-resource classes: given the image embedding from the original CLIP, we apply similarity searches with queries on each class, and randomly sample from the top images. For example, query ``\verb"a photo with dog"",
``\verb"a photo with cat"''
for the Pets class. 
% We also limit the image searching scope with practical rules. 
Since Booking.com has a substantial image dataset, we limit the image searching scope with practical rules. 
% For example, when searching for a certain facility (e.g., kitchen), we only search the images from the properties which have it. 

\subsubsection{Annotation Question Design}

We use AWS SageMaker Ground Truth and Mechanical Turk to collect annotations. This subsection describes how we get full annotation. 
% Below are the annotation collection designs, and how we get full annotation for each image. 
% on all 120 classes.

AWS provides quality control on the annotator pool, however, the quality could still variate with different tasks. We apply two steps to improve the quality: 1) Assign each annotation task to 5 annotators, and calculate the voting rate per image per label, to infer final ground truth. 2) Smart grouping: given one image, we can either ask 120 binary questions, or group labels into multi-choice questions. We annotate a ``golden ground truth" set, design different grouping strategies, and evaluate the annotators' performance. We find: binary questions get highest recall but lower precision, and cost the most; it's recommended to put mutually-exclusive labels in one group; never put many difficult labels together; it's advisable to make each question focusing on one scenario (e.g. Fireplace, Ceiling Fan and Air-conditioner in one group) and have generally 4-6 choices per group.   %  (max 6-7 choices)

\subsubsection{Consolidation and Dataset Results}
As stated above, we collect voting rate results from 5 annotators per image per class. Our evaluation on the golden ground truth set shows: with voting agreement threshold as 0.6, for the majority of classes, both precision and recall are high and near 0.95; for a few difficult classes, the annotation recall is typically higher than precision. Based on the annotation quality evaluation per class, we define customized agreement thresholds for each class. Besides that, we also apply the hierarchical mapping to improve the dataset quality: if one image is labeled as one class, then the class's parents are also marked as positive. We finally acquire more than 140K positive labels, and 7 million negative labels, on around 60K images. 

Even with all above practical improvements, the dataset still contains noise, which is a common challenge for real-world problems. We can tune the temperature of the loss function, and make it more tolerant to noise. The experimentation findings are shared in Section \ref{sec:temp_exp}.
% We can benefit from the temperature tuning of our loss function, to make it flexible for the noisy in-house dataset, and the findings are shared in section \ref{sec:results}.

\section{Experimental Setup} 
\label{sec:setup}
All of our experiments are performed on a computation instance equipped with 1 NVIDIA Tesla T4 Tensor Core GPU, 4 vCPU, and 16GB RAM. 
% and 125GB local storage. 
% We use AWS SageMaker to manage the experiments. For training, we use 1 NVIDIA Tesla T4 Tensor Core GPU, 4 vCPU, 16GB RAM and 125 local storage. 
The final best-performing model has the following settings: CLIP ViT-B/32 as the pretrained model; batch size of 64 images with full annotations; weight decay (on all weights that are not gains or biases) with coefficient 0.01, with AdamW optimizer \cite{weight_decay_adam} and learning rate 1e-5; positive class weight as 10 for all classes. Since we apply different image preprocessing from CLIP, use our own label-text input, and aim to get in-house image embeddings, we mainly experiment with unfreezing the image and text encoding backbones.  
% For other settings, we tune them according to the validation set performance, and this is the final best model's hyper-parameters: learning rate as 1e\-5, positive class weight as 10, weight decay coefficient as 0.02. 

We split the dataset into training set, validation set, and test set, with a ratio of 80/10/10 respectively. Since we have a class imbalanced multi-label dataset, 
we apply stratified sampling according to the lowest frequency label per image.  
% we apply stratified sampling on images, where each image is labeled as the lowest. 

To accelerate training and save memory, we apply the mixed-precision \cite{micikevicius2017mixed} strategy as CLIP does. During forward pass and backward propagation, half-precision is applied on the convolutional layers, linear layers, multi-head attention layers, text projection and image projection layers. For the weight update step, the full precision is applied to prevent underflow. The training takes roughly 40 minutes per epoch.

\subsection{Evaluation Metrics}

% In a multi-label classification scenario, the prediction can have three different results: fully correct, partially correct and fully incorrect. 
% Let’s assume the number of labels is $L$. For a given sample $x_i$ , the ground truth label $y_i$ and prediction probability $\hat{y_i}$ can both be represented as a vector with size $L$. 
Assume the number of labels is $L$, 
% For a given sample $x_i$ , the ground truth $y_i$ and prediction probability $\hat{y_i}$ can both be represented as a vector with size $L$. -> redundant 
and given $N$ image samples, the ground truth and the predictions can both be represented as a matrix with size $N \times L$. 
% , in which each cell has one binary value (since we are using binary ground truth). 
As \citet{eval_paper} mentioned, the evaluation is to 
% find approach to 
compare these two matrices and determine how close they are. 
% There are 3 commonly used comparison ways:
Comparisons can be made in three ways: column-by-column (also called label-based), row-by-row (example-based), or as a whole. 
% The column-based one is also named label-based, the row-based is also called example-based. 
% Since the column vector and row vector share the same representation(e.g. 0-1 vector), in theory, the same metric can be used in either way. 
We use below metrics as our main evaluation criterion:

\begin{itemize}
    \item 

Average Precision per class (label-based):

\begin{equation}
    AP_j =  \sum_{i=1}^{N} p_j(i)\Delta r_j(i) 
\end{equation}

where $p_j$ is the precision of class $j$, and $r_j$ is the recall of class $j$.  
It is equivalent to the area under the precision-recall curve per class. 

\item 
macro Mean Average Precision (aggregate on label-based):
\begin{equation}
    macro \: mAP = \frac{1}{L} \sum_{j=1}^{L} AP_j
\end{equation}

which is the unweighted average of AP across all classes.  

\item 
weighted Mean Average Precision (aggregate on label-based): 
\begin{equation}
    weighted \: mAP = \frac{1}{\sum_{j=1}^{L} {NP_j}} \sum_{j=1}^{L} AP_j \cdot NP_j
\end{equation}

where $NP_j$ is the number of positive samples of class $j$.
% , and $NP$ is the toatl number of positive samples

\item
Global Average Precision (global-based):
\begin{equation}
    GAP = \sum_{i=1}^{N \cdot L} p(i)\Delta r(i) 
\end{equation}

GAP (also called micro mAP \cite{yang1999evaluation}) is implemented as: collect predictions from all classes, and calculate the area under the global precision-recall curve. 
% https://www.kaggle.com/c/youtube8m/overview/evaluation 

\item
GAP@K: for each sample, take the top K predictions according to probability ranking, and calculate GAP on top of the collected predictions from all samples. 
\end{itemize}
\begin{figure}
    \centering
    \includegraphics[width=\columnwidth]{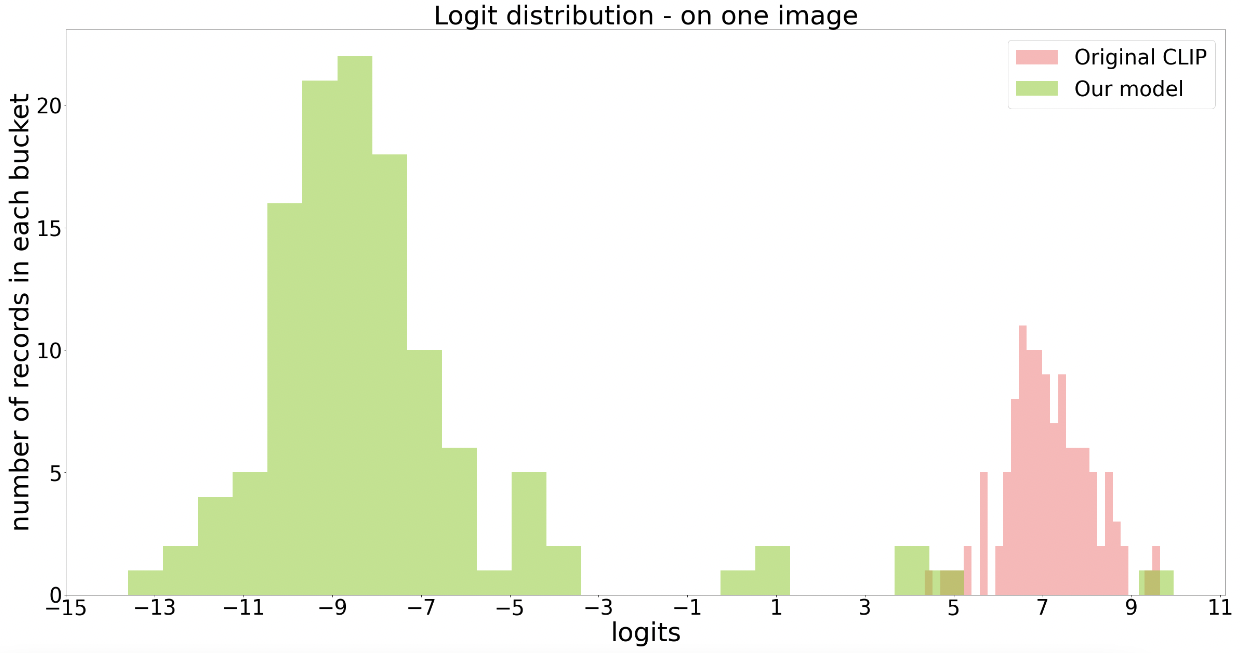}
        \caption{Output logits distribution - on one image, across 120 label-texts (``a photo with \{label name\}").}
    \label{fig:logit}
\end{figure}

\subsection{Baselines}
\label{sec:baselines}
We implement two baselines as follows: 
\begin{itemize}
    \item ASL \cite{ridnik2021asymmetric}: ASL authors provide a collection of high performance models trained 
    % with ASL 
    on various multi-label datasets \cite{asl_git}. We fine-tune the MS-COCO \cite{mscoco} based TResNet \cite{tresnet} large model, on our dataset, and apply hyperparameter tunes to get top-performing ASL model.
    \item Original CLIP \cite{radford2021learning}:  Given one image, CLIP was trained with the main objective as distinguishing one text from all others. Figure \ref{fig:logit} shows the output logits distribution from CLIP and MuMIC - on one image example. We observe that CLIP majority outputs are centered around value 7 for all classes, while MuMIC can distinguish between positive classes and negative ones. That is because CLIP was optimized for single-label with softmax-based loss, not for multi-label. 
    % It is clear that applying sigmoid on each logit would not provide valid predictions.  
    To generate multi-label predictions with CLIP, one approach is taking top $K$ output logits per image. However, it's not feasible to define a proper unified $K$. Instead, we apply the following approach: for each class, input a pair of text - ``\verb"a photo"'', ``\verb"a photo of {label name}"''; apply softmax on the two output logits, and use the second probability as the label prediction.
    
\end{itemize}

\section{Results and Development Findings}
\label{sec:results}
In this section, we explain the final model performance and other experimental findings. 
% In this section, we show the final model performance, the ablation study results on temperature tuning, the prompt engineering on label-text input, the visualization on image embedding, and the zero-shot prediction. 
\subsection{Final Model Performance}

% To generate multi-label predictions, one way is to take the top K predictions per image, however, it is difficult to select K

% \begin{figure}[H]
%     \includegraphics[width=0.7\columnwidth]{./figures/35217249.jpeg}
%         \caption{One image example}
%     \label{fig:devtypes}
%     \centering
% \end{figure}  

\begin{table}  % was [H]
\begin{tabular}{
>{\columncolor[HTML]{FFFFFF}}l 
>{\columncolor[HTML]{FFFFFF}}r 
>{\columncolor[HTML]{FFFFFF}}r 
>{\columncolor[HTML]{FFFFFF}}r 
>{\columncolor[HTML]{FFFFFF}}r }
\hline
method & \multicolumn{1}{l}{\cellcolor[HTML]{FFFFFF}GAP} & \multicolumn{1}{l}{\cellcolor[HTML]{FFFFFF}GAP@10} & \multicolumn{1}{l}{\cellcolor[HTML]{FFFFFF}\begin{tabular}[c]{@{}l@{}}macro \\ mAP\end{tabular}} & \multicolumn{1}{l}{\cellcolor[HTML]{FFFFFF}\begin{tabular}[c]{@{}l@{}}weighted \\ mAP\end{tabular}} \\ \hline
CLIP   & 32.7                                            & 44.5                                               & 56.7                                                                                             & 56.1                                                                                               \\
ASL    & 67.7                                            & 77.3                                               & 60.8                                                                                             & 68.9                                                                                                \\
MuMIC  & {\color[HTML]{1A1C1F} \textbf{83.8}}            & {\color[HTML]{1A1C1F} \textbf{85.6}}               & {\color[HTML]{1A1C1F} \textbf{74.7}}                                                             & \textbf{79.5}                                                                                       \\ \hline
\end{tabular}
\caption{Performance comparison of MuMIC against 2 baselines, on the test set, 120 classes. All metrics are in \%.} 
\label{tab:perf}
\end{table}

\begin{figure}
    \centering
    \includegraphics[width=\columnwidth]{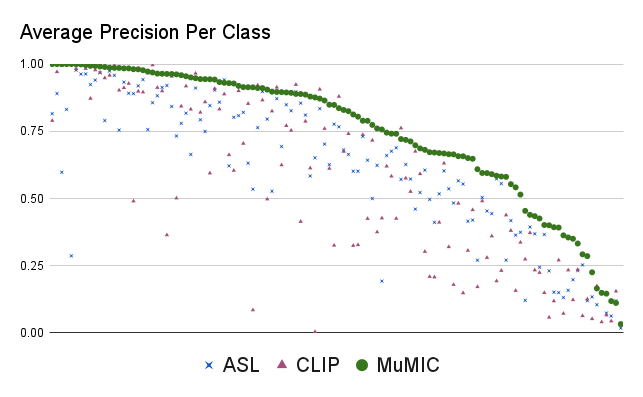}
        \caption{AP on each class, on the test set. The x axis represents the 120 classes (decreasing order on MuMIC AP). }
\label{fig:ap_120}
\end{figure}

Table \ref{tab:perf} and Figure \ref{fig:ap_120} show how the final MuMIC model outperforms the two baselines. 
From Figure \ref{fig:ap_120}, we observe that MuMIC outperforms the other 2 baselines across large majority of the labels. Furthermore, we see a few classes for which none of the compared methods achieve satisfactory results. 
Our investigation shows two main reasons: these classes are more noisy since they are challenging and not easily distinguished even by a human (e.g. Toaster, Patio); 
% despite an additional enrichment step in annotation collection, 
they are still too low frequent.  
% and have very small resolution in our images. -> already mentioned in first point (difficult for human)
% Having said that, we still include them in the evaluation to get an overview on all predefined classes. -> redundancy words 
Table 2 shows AP scores on some classes as examples. For general class like Bed or Food, all methods show acceptable performance. For more travel domain related classes, model training is necessary. For example, Animal in travel does not include pets, and there are various types of Lobby, where we observe the original CLIP performance is not enough. 
% We also find some low resource classes can reach high performance, like the Winter, which has 1
% Figure \ref{fig:pred_3} shows a snippet of MuMIC prediction examples. 

% AP \\ per \\ class
\begin{table}
\begin{tabular}{lrrrrrr}
\hline
method & \multicolumn{1}{l}{Golf} & \multicolumn{1}{l}{Bed} & \multicolumn{1}{l}{\begin{tabular}[c]{@{}l@{}}Win-\\ ter\end{tabular}} & \multicolumn{1}{l}{Food} & \multicolumn{1}{l}{\begin{tabular}[c]{@{}l@{}}Lob- \\ by \end{tabular}} & \multicolumn{1}{l}{\begin{tabular}[c]{@{}l@{}}Animal \\ (not \\ pets)\end{tabular}} \\ \hline
CLIP   & 94.8                     & 95.8                    & 91.2                                                                   & 90.0                     & 50.1                                                                               & 49.0                                                                                \\
ASL    & 79.0                     & 97.4                    & 93.3                                                                   & 91.3                     & 73.3                                                                               & 89.0                                                                                \\
MuMIC  & \textbf{98.9}            & \textbf{98.6}           & \textbf{98.4}                                                          & \textbf{96.4}            & \textbf{96.2}                                                                      & \textbf{98.1}                                                                       \\ \hline
\end{tabular}
\label{tab:ap_per_class}
\caption{Average Precision (AP) on some classes - MuMIC against 2 baselines on the test set. All metrics are in \%.}
\end{table}

Regarding the training time efficiency, the ASL training time per epoch is similar to MuMIC (40 minutes as Section \ref{sec:setup} shows), but ASL takes around 18 epochs to saturate, while MuMIC only needs around 3 epochs. 
% - the validation performance starts saturation at around epoch 18, while MuMIC starts saturation from epoch 2, and reaches the final performance at the end of epoch 3. 
That indicates MuMIC training efficiency is boosted by the large-scale pretrained CLIP, and the hardness-aware loss function.  
% (on 400 million dataset). 
% For inference, in real-life cases of multi-label, the CLIP prediction time complexity is twice than MuMIC, since CLIP requires a pair of text input for each class. 
Regarding the inference time, CLIP's text encoding time is twice that of MuMIC, since CLIP requires a pair of text inputs per class. 

\subsection{Temperature Factor Selection}
\label{sec:temp_exp}
As described in Section \ref{sec:approach}, the temperature $\tau$ is an important factor. 
CLIP initializes the contrastive loss temperature as 0.07 (empirical value from \citet{understand_cl}), and clips the value at around 0.01 to prevent training instability. We also decide to initialize temperature, make it a parameter for the model to learn with capping value, instead of freezing it as a hyperparameter. Since we have a real-world noisy dataset, and are fine-tuning on top of pretrained models, we find it is necessary to search the best initialization temperature for the tempered sigmoid. 

\begin{figure}  % was [H] - before add section deploy
    \centering
    \includegraphics[width=0.95\columnwidth]{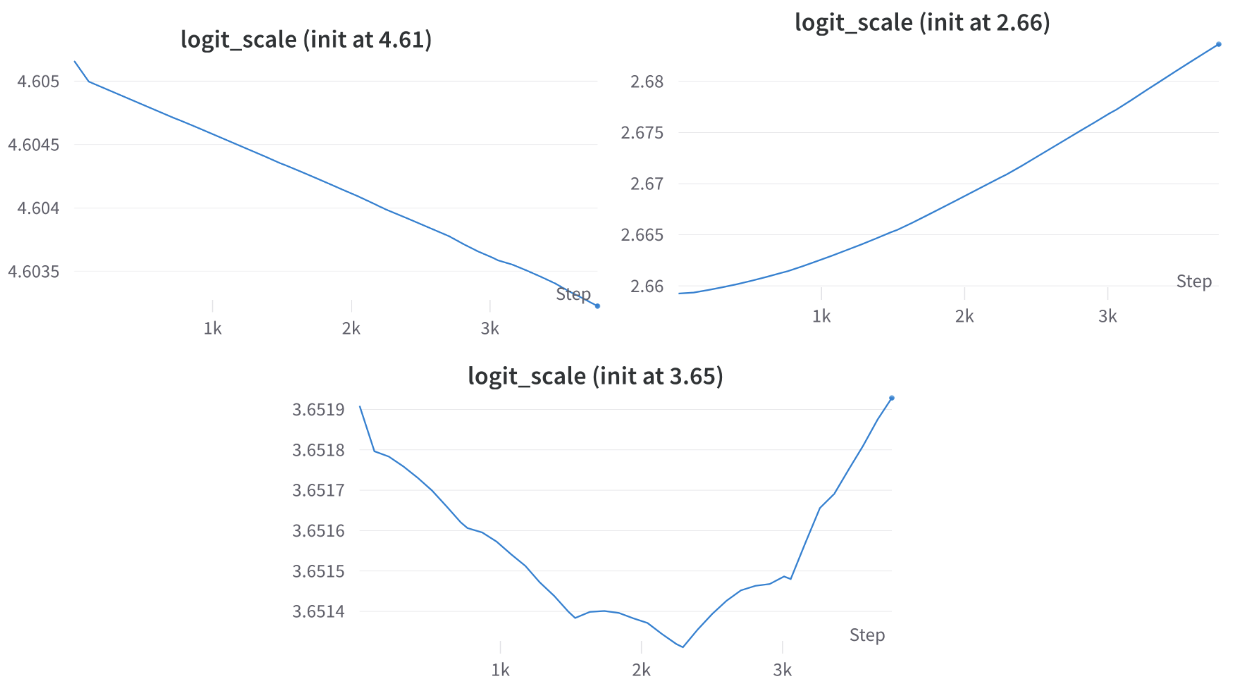}
        \caption{Param $logit\_scale$ learning curve examples}
    \label{fig:logit_all}
\end{figure}

% the full boarder version: 
% \begin{table*}
% \centering
% \begin{tabular}{|l|r|r|r|r|r|r|r|r|r|r|r|}
% \hline
% \begin{tabular}[c]{@{}l@{}}logit\_scale \\ init value\end{tabular} & 0.70 & 0.91 & 2.00 & 2.58 & 2.66 & 2.93 & 3.35 & 3.65 & 4.61 & 5.24 & 5.75 \\ \hline
% \begin{tabular}[c]{@{}l@{}}macro mAP\end{tabular}                  & 17.5 & 22.8 & 65.9 & 72.2 & 72.6 & 73.7 & 73.9 & 74.1 & 73.1 & 73.1 & 72.6 \\ \hline
% GAP@10                                                                & 56.1 & 62.1 & 78.5 & 83.9 & 83.9 & 84.7 & 85   & 85.5 & 84.4 & 84.4 & 84.1 \\ \hline
% \end{tabular}
% \label{tab:temp}
% \caption{Validation set performance - with different temperature initialization value. }
% \end{table*} 

\begin{table*}
\centering
\begin{tabular}{lrrrrrrrrrrrr}
\hline
\begin{tabular}[c]{@{}l@{}}logit\_scale \\ init value\end{tabular} & 0.703 & 0.913 & 2.005 & 2.578 & 2.659 & 2.927 & 3.352 & \textbf{3.652} & 4.189 & 4.605 & 5.244 & 5.751 \\ \hline
macro mAP                                                          & 17.5  & 22.8  & 65.9  & 72.2  & 72.6  & 73.7  & 73.9  & \textbf{74.1}  & 73.9 & 73.1  & 73.1  & 72.6  \\
GAP@10                                                             & 56.1  & 62.1  & 78.5  & 83.9  & 83.9  & 84.7  & 85.0    & \textbf{85.4} & 85.0  & 84.4  & 84.4  & 84.1  \\ \hline
\end{tabular}
\label{tab:temp}
\caption{Validation set performance - with different temperature initialization values }
\end{table*}

We perform Bayesian optimization \cite{bayesian1978} search to find a proper setting. Table 3 % \ref{tab:temp} 
shows the validation set performance with 
various initialization values.  
Our best $logit\_scale$ initialization value is $3.652$, which outperforms the cases when initializing $\tau$ either as 0.07 ($logit\_scale$ $2.659$), or as the pretrained CLIP $\tau$ value ($logit\_scale$ $4.605$).   
% - equivalent to load pre-trained CLIP when we don't re-initialize (continue from the pre-trained value, which is around temperature 0.01),
% As a network parameter, model updates the $logit\_scale$ step by step. 
Figure \ref{fig:logit_all} shows how the network learns to optimize $logit\_scale$ during training steps. 

% Our experiment also show that the validation set performance saturate at around step 2k (at the end of epoch 3)

% Besides that Figure, from validation set performance curve, we observe the convergence at the end of epoch 3 (step around 2K). -> a bit confusing. 

From both Table 3 and Figure \ref{fig:logit_all}, we observe it is important to start $\tau$ from a proper value. If the temperature is too small, the strict penalties on hard samples might make the model too sensitive to noise like wrong annotations. 
If the temperature is too big, then the penalties on hard samples could be not enough and limit the model's learning ability. 
% and leads to lower performance. 
In addition, we run an experiment with normal sigmoid-based BCE loss (freeze $\tau$ at value 1), 
% ($\tau$ is set as a static hyperparameter with value 1), 
the validation set can reach a maximum macro mAP score of 43.3\% and GAP@10 score of 38.2\%, which are far lower than MuMIC performance. 
% in Table 3. 
% That is expected as explained in section \ref{sec:bce_loss}.    
% Last sentence can remove if no space left. 
% \begin{figure}
%     \includegraphics[width=0.8\columnwidth]{./figures/val_perf.png}
%         \caption{Validation performance - with init value at 3.652}
%     \label{fig:val_perf}
%     \centering
% \end{figure}   
% \begin{figure}  % temp remove
%     \centering
%     \includegraphics[width=0.8\columnwidth]{./figures/pred_3_sc.png}
%         \caption{MuMIC prediction examples}
% \label{fig:pred_3}
% \end{figure} 

\subsection{Enrich Class Context with Class Description}

% \guyn{As described in section Approach, we experimented with either using "a photo with {label name}", or "a photo with {label description}" as the input text representing a label. This led to label description being used across multiple labels that benefited from it, as shown in Figure \ref{fig:desc}.} -> adjusted below
\begin{figure}
    \centering
    \includegraphics[width=0.9\columnwidth]{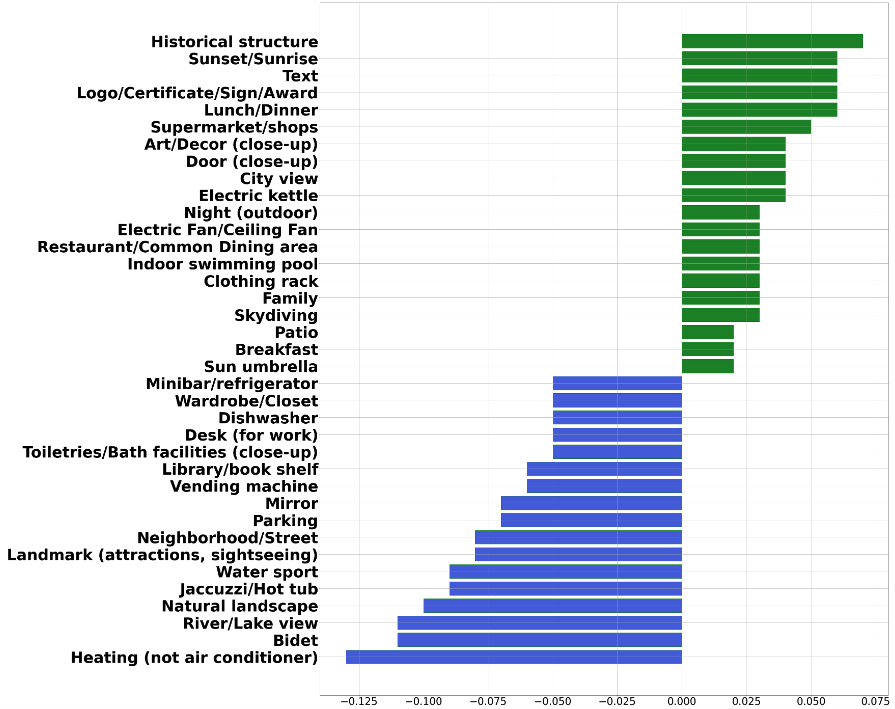}
        \caption{Validation set AP gap examples. The x axis is the AP gap: using class description minus using class name.}
    \label{fig:desc}
\end{figure} 

As described in Section \ref{sec:approach}, we experiment with either using ``\verb"a photo with {label name}"'', or ``\verb"a photo with {label description}"'' as the input text representing a label. This leads to label description being used across a subset of labels that benefit from it, as the upper side examples shown in Figure \ref{fig:desc}. 
More convenient labels like Parking and Mirror prefer concise class names rather than descriptions, in comparison to labels like Art/decor, Historical structure where descriptions improve the label-text embedding and the performance. 
% -> exceed page after adding. reader can find from Figure themselves
% that we see adding a description improve the label-text embedding and the label performance

\subsection{Image Embedding}

\begin{figure}[H]
    \centering
    \includegraphics[width=0.38\textwidth]{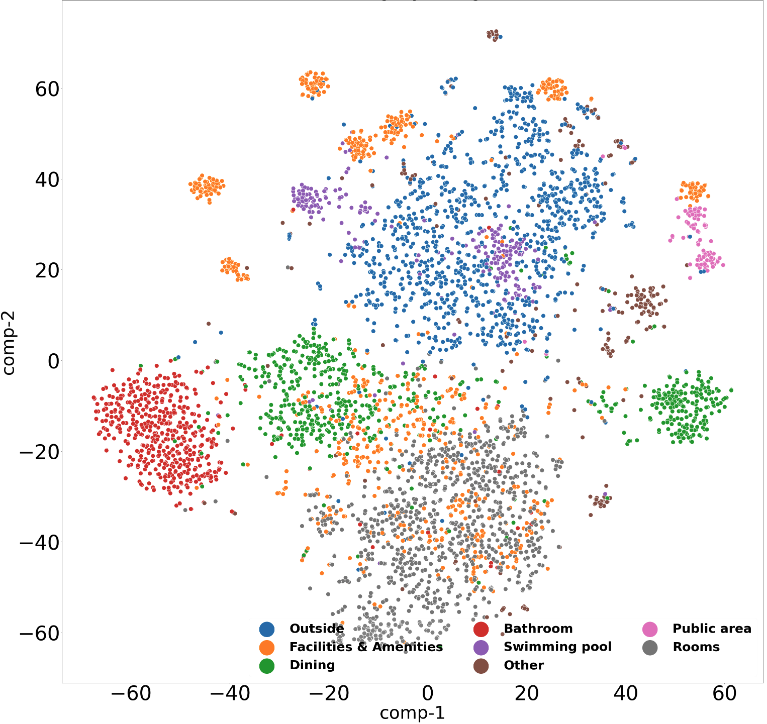}
        \caption{t-SNE on category - original CLIP}
    \label{fig:tsne_ori_cat}
\end{figure} 

\begin{figure}[H] % was [H]
    \centering
    \includegraphics[width=0.38\textwidth]{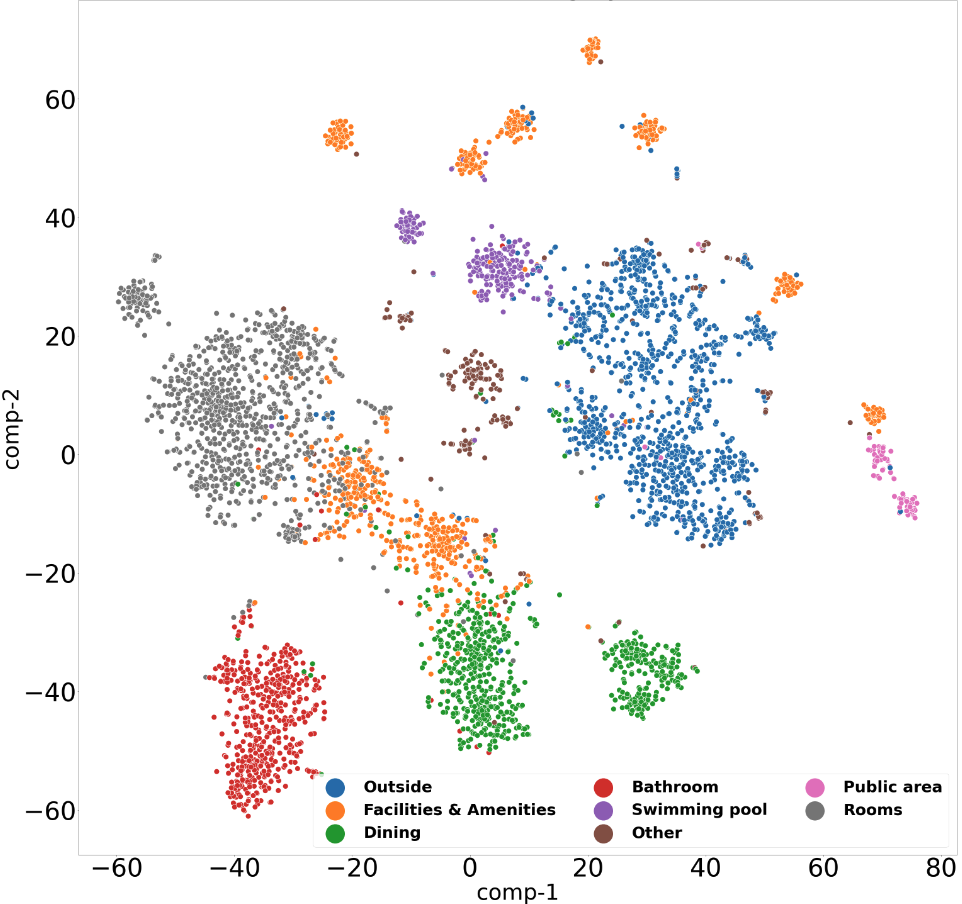}
        \caption{t-SNE on category - MuMIC}
    \label{fig:tsne_our_cat}
\end{figure} 

MuMIC generates travel domain image embeddings which can be applied to downstream machine learning applications (e.g. gallery embedding to find similar hotels). Figure \ref{fig:tsne_ori_cat} and \ref{fig:tsne_our_cat} display the t-SNE \cite{van2008visualizing} visualizations of image embeddings on the test set, colored at the category level. When an image has multiple labels, we color the point with the lowest frequent class's category, to acquire more samples for low resource categories. From the two figures, MuMIC has better ability to enforce differentiation among different categories. We also analyze t-SNE distributions on the class level, and reach similar conclusions.

% \begin{figure}[H]
%     \includegraphics[width=\columnwidth]{./figures/tsne_ori_32tag_color.png}
%         \caption{tSNE on 32 most important tags - original CLIP}
%     \label{fig:devtypes}
%     \centering
% \end{figure}  

% \begin{figure}[H]
%     \includegraphics[width=\columnwidth]{./figures/tsne_32tag_color.png}
%         \caption{tSNE on 32 most important tags}
%     \label{fig:devtypes}
%     \centering
% \end{figure}  

% (CLIP mention weight decay L2 penalty tends to select a big penalty on the weight - limitation on few-shot). 

\subsection{Zero-shot Learning}

With multimodal learning, we expect semantic information to propagate from seen classes to the unseen ones. As the predefined 120 classes covered main travel topics, MuMIC can generalize well to unseen travel domain labels. 
% by acquiring knowledge from learnt labels, 
% and perform better than CLIP.  

\begin{figure}
    \centering
    \includegraphics[width=0.8\columnwidth]{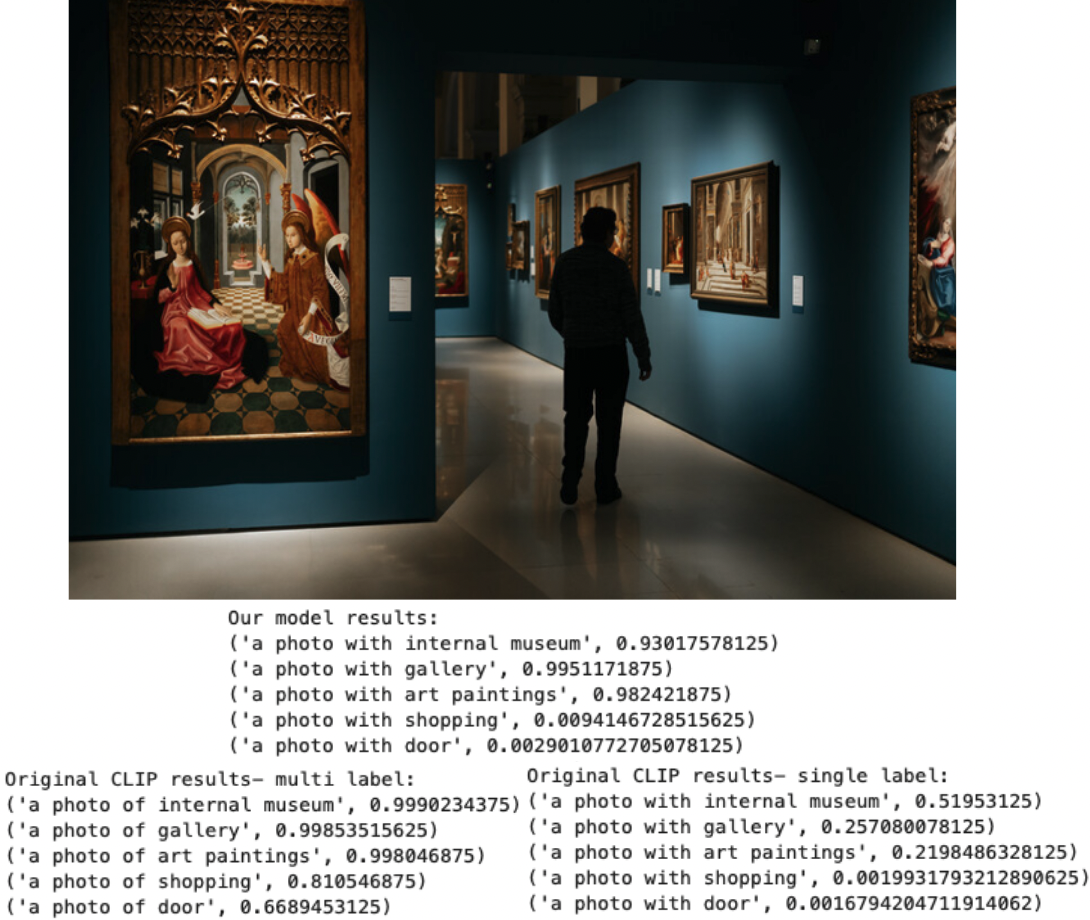}
        \caption{Zero-shot Prediction examples - Museums}
    \label{fig:museum}
\end{figure}

\begin{figure}
    \centering
    \includegraphics[width=0.8\columnwidth]{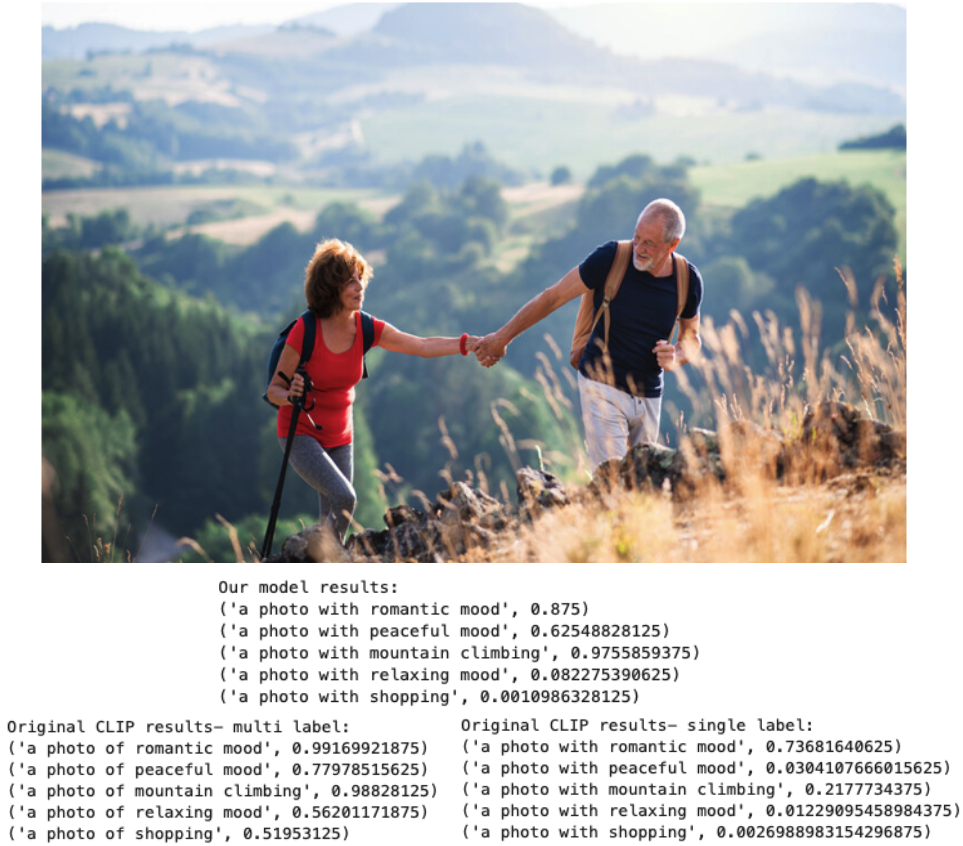}
        \caption{Zero-shot Prediction examples - Trip types}
    \label{fig:romantic}
\end{figure}

Figure \ref{fig:museum} 
and \ref{fig:romantic} show zero-shot prediction examples from MuMIC and CLIP. 
% , where MuMIC gets better predictions. 
% shows one image example of the prediction scores from MuMIC and CLIP, on 5 new classes. 
% Figure \ref{fig:museum} shows zero-shot on 5 new classes. 
For MuMIC, we apply sigmoids on the 5 output logits to get prediction scores. 
For CLIP, we apply two approaches: multi-label as Section \ref{sec:baselines} described; and single-label via softmax on the 5 output logits. MuMIC gets high scores on correct labels (internal museum, mountain climbing etc.), and low scores on wrong labels (shopping, door etc.), indicating good generalization. Comparing with MuMIC, CLIP multi-label approach gets quite high scores on wrong labels. CLIP single-label approach has the top K selection problem as Section 4.2 explained. 
% , brings difficulties to consume the scores.  
% We see MuMIC gets higher probabilities for the right classes (museum, art paintings and gallery) and lower probabilities to the wrong classes, which indicates MuMIC is more accurate.
% Besides this Figure, we also observe that MuMIC performs better in more examples. 

\section{Application Examples}
\label{sec:application}

% Applications in the traveling domain that can benefit from MuMIC are:
The final MuMIC model supports multiple downstream use cases. For each application, online A/B test \cite{Fabijan2018ExperimentationGE} is required to be conducted on real traffic of Booking.com and typically lasts for several weeks. Below are application examples that can benefit from MuMIC: 
% In this section, we describe the main use case examples. 
% which can drive up booking conversion rate. 
% Examples of applications in the travel domain that can benefit from MuMIC are as follows, which drive up booking conversion rates: 
% Below are application examples that can benefit from MuMIC: 

% \begin{itemize}
% \item

\begin{figure}
    \centering
    \includegraphics[width=0.8\columnwidth]{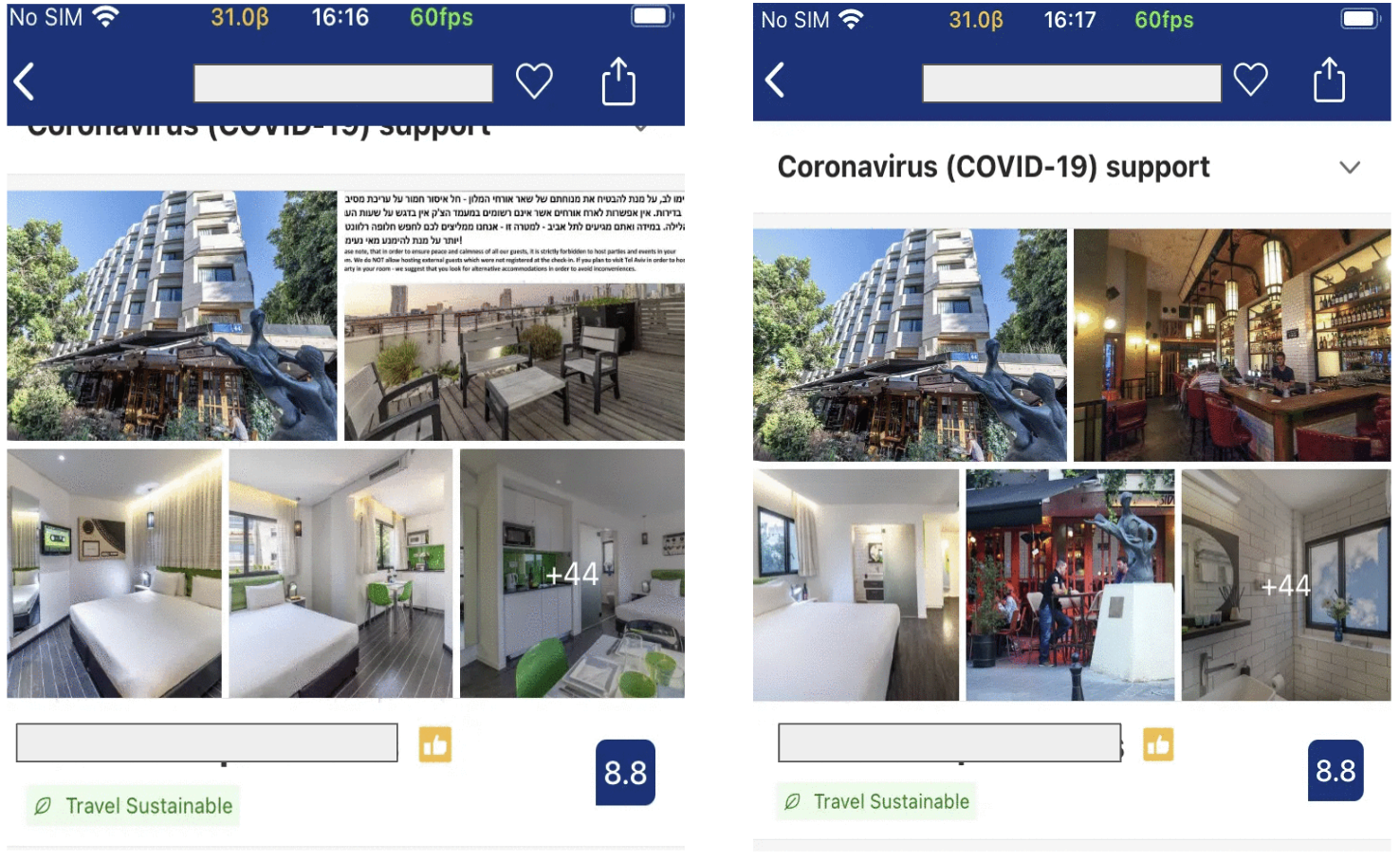}
    \caption{Gallery subset display}
    \label{fig:gallery}

\end{figure}

\textbf {Main gallery subset selection} By default, the gallery subset/preview on the property page 
% (5 images on the Booking.com Apps) 
contains the first images uploaded. MuMIC can be applied to select more representative subset to cover different aspects - building, room, bathroom, dining area, etc. 
% , which sometimes does not provide the full profile of the property and may include repeating photos. 
% MuMIC can be used for choosing photos from different categories. 
% , to have a diversity subset of images.
Figure \ref{fig:gallery} shows one example: the left side is before applying the model, where we see duplicate bedroom images, and an image with long text; the right side is the new design, shows better gallery quality. This A/B test results in +0.5\% more users progressing to the next step of the booking funnel, indicates better user engagement. 
% where images from different categories are displayed in the main subset. 

% \item 
% \textbf{Gallery embedding}: MuMIC image embeddings can be applied to generate gallery embeddings for other machine learning models (e.g. find similar hotels). 
% macapplications like similar hotel searching etc. 

% \item -> a bit duplicate use case 
% \textbf{Content completion}: MuMIC can suggest labels (i.e, oven, microwave) on the room level images in order to remind property managers to annotate them, 
% % on facilities they didn't annotate, 
% which will results in an improved user search experience. 
% % increase the match between user search and partner property. 
% % In addition, suggesting relevant labels to partners in the process of their gallery upload.

% \item -> TODO: consider comment out if no space. 
\textbf{Content selection}
% MuMIC can suggest best image candidates for themed social campaigns and promotions, native ads, etc. based on the label predictions. 
a) Native ads - the image classification results are applied in a combination with an image quality model in order to select the best images for native ads. The A/B test results in +5\% more clicks on ads. 
b) Destination recommendation - some destinations have no image on their recommendation cards. With image classification results on outdoor classes (natural landscape, street, sightseeing, etc.), we are able to re-purpose gallery photos and use them for supplementing missing destination images. The A/B test results in +4.3\% more clicks on the cards. 

\textbf{Content validation and enrichment} - identifying mismatches between the facilities reported by the property owners and the ones existing in property’s photos, and enabling owners to take actions. This is currently in production on Booking.com partner website, with property owners adopting 96\% of the recommended gaps. When customers filter for a certain facility, for example swimming pool, they are able to see and book more properties with swimming pools. 

% \textbf{Gallery personalization}: 
% % TODO: maybe delete if no space 
% suggesting gallery image ranking based on the user segment.  
% % by matching relevant images. 
% For example, families will be interested in images containing aqua park, game room, and children; while business travellers will be interested in room amenities, conference room, and breakfast. 

% \end{itemize} 

\section{Deployment and Maintenance}
The final MuMIC model is served with Amazon SageMaker, and deployed on Booking.com Content Intelligence Platform (CIP) - a stream processing platform based on Apache Flink, that consumes real-time events (e.g. requests on image prediction) from Kafka topics, and generates model-based results. The final product provides both real-time predictions services, and backfilling services.  

% or performs object operations. 
% The model is deployed .
% Besides providing real-time model prediction services, we also perform backfilling.
The backfilling jobs run MuMIC on Booking.com image datasets, 
% : we produce events into a dedicated Kafka topic, with the image metadata. 
store the predictions, image embeddings and metadata to databases. Thus, consumers can retrieve results from a central place, instead of sending and managing requests on the overlapping images per use case. 
To save storage space, we decide to only save the label predictions above certain thresholds. We perform thresholds selection per class on the validation and test set, by analyzing the percentage of data we can drop 
% (the ratio between the number of samples below the threshold, and the total samples), 
and the maximum recall still remain. We finally drop around 87\% of data while keep the recall close to 1. 

The model is served on GPU instances instead of CPU, since our analysis shows that to predict on same amount of images, the cost on using GPU is around 3 times less than CPU. It's also worth mentioning the model inference endpoints support batch predictions, which is shown to be roughly 5 times faster than predicting images one by one. The above optimizations match our expectations, since the MuMIC architecture is highly parallelizable because of the transformers backbones. 
% we need to decide reasonable thresholds for each class, which guarantee a high recall, as well as drop the majority of false positives. 

To achieve system robustness and scalability, we apply auto-scaling, which leads to a better handling on peak times, and a lower cost on off-peak times. In addition, model serving metrics (e.g. throughput, inference time) are monitored using in-house dashboards and AWS CloudWatch. 
% Besides, the backfilling stored image embeddings can be reused for zero-shot predictions. 
% When the traffic is small, the model consumes lower resource and lower cost. 

% probability storage: 350 million * ((120 float * 4byte * 8) + (1 long int * 8byte * 8) 
% + (1 int * 4byte * 8) ) = 1283GB 

\section{Conclusion}

Multi-label image classification is a crucial task for many domains. 
% In this paper we show the relevancy of this task to the traveling domain. 
In this paper, we present MuMIC, a multimodal approach for multi-label image classification based on contrastively learnt CLIP model. 
Our main novelty lies in the creation of an in-house dataset for the travel industry, and application of supervised multimodal learning with tempered-sigmoid based BCE loss. 
We perform ablation studies and show the impact of different choices, including temperature tuning, and class context enrichment with class descriptions.

Our development, performed on a real-world Booking.com dataset, demonstrates that MuMIC is a practical framework that outperforms SOTA approaches in both classification performance and training efficiency. With MuMIC framework, the model also learns high quality in-domain image embeddings, 
and acquires zero-shot learning abilities on unseen classes without additional training. 
% MuMIC trained model is practical and useful for new classification tasks (especially within same domain) without additional training.

For future work, we suggest improving the low performance labels, 
by refining class definitions and representations, reducing annotation noise, and assigning higher loss weights. Few-shot learning for new classes on top of MuMIC embeddings could also be a promising direction. 

\section*{Acknowledgments}
This work is supported by Booking.com. We would like to thank David Konopnicki, Monika Wysoczanska, Nerie Ohana, Dima Goldenberg, and Amit Beka for paper review. 
% reviewing the whole paper and providing suggestions. 

% Future work? 
% Another idea is to enable soft labels by using voting rate as ground truth. Since the difficulty for annotate each tag is different, if there is a plan to try it, the voting rate score also need to be adjusted properly. 
% rotation: not there. [though it is rare to see] [can be next version]. 
% class weight more on low resource: but after several epochs, it will focus on lifting them too. 

% Use \bibliography{yourbibfile} instead or the References section will not appear in your paper
\bibstyle{aaai23}
\bibliography{aaai23}% without the .bib extension
\end{document}